# Formal Context Reduction in Deriving Concept Hierarchies from Corpora Using Adaptive Evolutionary Clustering Algorithm

**Bryar A. Hassan[1,2], Tarik A. Rashid[3], Seyedali Mirjalili[4,5]**

[1] Department of Computer Networks, Technical College of Informatics, Sulaimani Polytechnic University, Sulaimani, Iraq

[2] Kurdistan Institution for Strategic Studies and Scientific Research, Sulaimani, Iraq

[3] Computer Science and Engineering Department, University of Kurdistan Hewler, Erbil, Iraq

[4] Centre for Artificial Intelligence Research and Optimisation, Torrens University, Australia.

[5] Yonsei Frontier Lab, Yonsei University, Seoul, Korea.

**Corresponding Email:** bryar.hassan@kissr.edu.krd

**Abstract**

It is beneficial to automate the process of deriving concept hierarchies from corpora since a manual construction of concept hierarchies is typically a time consuming and resource-intensive process. As such, the overall process of learning concept hierarchies from corpora encompasses a set of steps: parsing the text into sentences, splitting the sentences and then tokenised it. After the lemmatisation step, the pairs are extracted using formal context analysis (FCA). However, there might be some uninteresting and erroneous pairs in the formal context. Generating formal context may lead to a time-consuming process, so formal context size reduction is require to remove uninterested and erroneous pairs, taking less time to extract the concept lattice and concept hierarchies accordingly. In this premise, this study aims to propose two frameworks: i) A framework to review the current process of deriving concept hierarchies from corpus utilising formal concept analysis (FCA); ii) A framework to decrease the formal context's ambiguity of the first framework using an adaptive version of evolutionary clustering algorithm (ECA*). Experiments are conducted by applying 385 samples corpora from Wikipedia on the two frameworks to examine the reducing size of formal context, which leads to yield concept lattice and concept hierarchy. The resulting lattice of formal context is evaluated to the standad one using concept lattice-invariants. Accordingly, the homomorphic between the two lattices preserves the quality of resulting concept hierarchies by 89% in contrast to the basic ones, and the reduced concept lattice inherits the structural relation of the standard one. The adaptive ECA* is examined against its four counterpart baseline algorithms (Fuzzy K-means, JBOS approach, AddIntent algorithm, and FastAddExtent) to measure the execution time on random datasets with different densities (fill ratios). The results show that adaptive ECA* performs concept lattice faster than other mentioned competitive techniques in different fill ratios.

**Keywords**

Concept hierarchies, formal context reduction, concept lattice reduction, adaptive ECA*, FCA, WordNet.

## 1. Introduction

The Semantic Web is an extended web of machine-readable data, which provides a program to process data via machine directly or indirectly [1]. As an expansion of the latest Web, the Semantic Web can add meaning to the World Wide Web content and thus support automated services on the basis os semantic representations. Meanwhile, the Semantic Web depends on structured ontologies to organize the underlying data and provide a detailed and portable interpretation of computing machines [2]. Ontologies, as an essential part of the Semantic

Web, are commonly used in Information Systems. Also, the proliferation of ontologies demands that ontology development should be derived quickly and efficiently to bring about the Semantic Web's success [1]. Nonetheless, manually constructing ontologies is still a time-consuming and tedious process, with the bottleneck of information acquisition, time-consuming growth, maintenance difficulty, and the integration of different ontologies for different applications and domains being the key issues. Ontology learning is a solution to the bottleneck of gaining knowledge and creating ontologies in a timely manner. Ontology learning is a subtask of knowledge extraction whose aim is to create ontologies by semi-automatically or automatically extracting related concepts and relationships from a corpus or other data sets. The current approach for automatic attainment of concepts and concept hierarchies form corpus is introduced by [3]. This approach is based on FCA to construct a concept lattice that can be converted to a particular type of partial order that constitutes a hierarchy of concepts. On this basis, the main contributions of this research are proposing two frameworks for deriving concept hierarchies automatically from free corpora as follows:

- The first framework includes a set of steps to construct concept hierarchies from the text. One of the primary steps of this framework is how to extract the word pairs or phrase dependencies, especially the necessary and interested pairs, because not all pairs are correct or interesting. Based on some statistical measure, the pairs are then weighted, and the word pairs over a defined threshold are converted into a formal context according to which FCA is applied. Lastly, the concept hierarchies are produced from the concept lattice.
- The second framework uses the same steps of the former, but it excludes erroneous and uninterested pairs from the formal context, resulting in a smaller formal context and less time spent deriving the concept lattice. To do so, the adaptive version ECA* is applied to the formal context, thereby producing a reduced concept lattice and concept hierarchy as well. The produced concept lattices from the frameworks are evaluated to examine how they are isomorphic to each other.

One of the evolutionary clustering algorithms proposed recently in [4] is called ECA*. In this approach, various techniques are combined, such as classification of the social classes; percentiles and quartiles; optimisation algorithm operators, and K-means algorithm. One significant component of ECA* is the deployment of the Backtracking Search Optimisation Algorithm (BSA). This method is considered as one of the popular evolutionary techniques that outperformed its counterpart algorithms due to its recombined strategy of mutation and crossover [5, 6]. Meanwhile, one of the most significant parts of ECA* is the mut-over operator [4]. The core strength of ECA* is the exploitation of the mut-over operator as a recombination approach (mutation and crossover), which is extracted from BSA. The rationale behind conducting this study is primarily to deduct the formal context size to derive a concise and meaningful concept hierarchy from text corpora. Significantly, the size reduction of formal context leads to less complexity of concept lattice and concept hierarchy as well. This premise has three advantages: (1) Producing concept hierarchies in this way may lead to a time-consuming process; (2) Using mut-over operator of an adaptive ECA* in this process can tune to a meaningful result of concept hierarchies; (3) decreasing the size of concept hierarchies results in eliminating erroneous and uninterested information.

The rest of this research is organized in the following structure: Section 2, reviews the past work related to deriving concept hierarchies from corpora and reduction of concept lattices in the literature. Section 3, presents a proposed framework to decrease the formal context size that can produce concept lattice for deriving concept hierarchies from free text using adaptive ECA*. In Section 4, an experiment is carried out on 385 datasets to test the reduced concept lattices with the original ones, followed by evaluating methods the results achieved from the experiment. Section 5 presents an experimental comparison between adaptive ECA* and its counterpart algorithms. Finally, the concluding remarks are addressed, with the reflection of this work's weaknesses to clarify how this study will help scholars, accompanied by suggestions for potential research.

**2. Related literal work**

Generating automatic ontologies is one of the on-going researches in the Semantic Web area. Ontology learning is considered as a resource-demanding and time-consuming task - with the involvement of domain experts to generate the ontology manually [7]. As a result, endorsing this method partially or entirely for semi-automatically or ultimately constructing ontology would be helpful. Ontology learning can be identified as a subtask of information extraction to semi-automatically derive relations and relevant concepts from text corpora or other types of data sources to construct an ontology.

Over the last decade, a variety of techniques from the fields of machine learning, natural language processing, knowledge representation, information retrieval and data mining have helped to improve ontology growth [7]. Data mining, machine learning, and knowledge retrieval are computational methods that can be used to retrieve particular domain words, definitions, and connections. Natural language processing, on the other hand, plays a key role in almost every stage of the ontology learning layer cake by offering linguistic techniques. FCA is one of the interesting statistical techniques used to construct concept hierarchies. This approach depends on the cocnept that objects are related to their corresponding attributes property. It takes a matrix attribute of an object as input and determines all the natural attributes and object clusters together. It yields a lattice in the shape of a hierarchy, with definitions and attributes. The concept of the general use of FCA is undoubtedly not new. In [8], The FCA's potential applications include linguistic structure analysis, lexical semantics, and lexical tuning, to name a few. Recently, [3] proposed a novel approach to automatically acquiring concept hierarchies from domain-specific texts focused on FCA. The method was also compared to a hierarchical agglomerative clustering algorithm and Bi-Section-K-Means, and it was discovered that it worked better on two datasets.

Various information measures were also tested in order to decide the value of an attribute/object pair, and it was found that the conditional probability works well in contrast to other more complicated information measures. The overall process of automatically deriving definition hierarchies from text corpora was depicted in the same analysis. A collection of steps are used in the overall process of automating concept hierarchies using Systematic Concept Analysis. To build a parse tree, the text is first labelled as part-of-speech (POS) and then parsed for each sentence. The parse trees are then used to extract dependencies between verb/object, verb/subject, and verb/prepositional expression. In particular, pairs containing the verb and the head of the object, subject, or prepositional sentence are extracted. After that, the verb and its head are lemmatised and moved to their base form. The method of extracting word pairs from text corpora is known as NLP components, as shown in Figure 1.

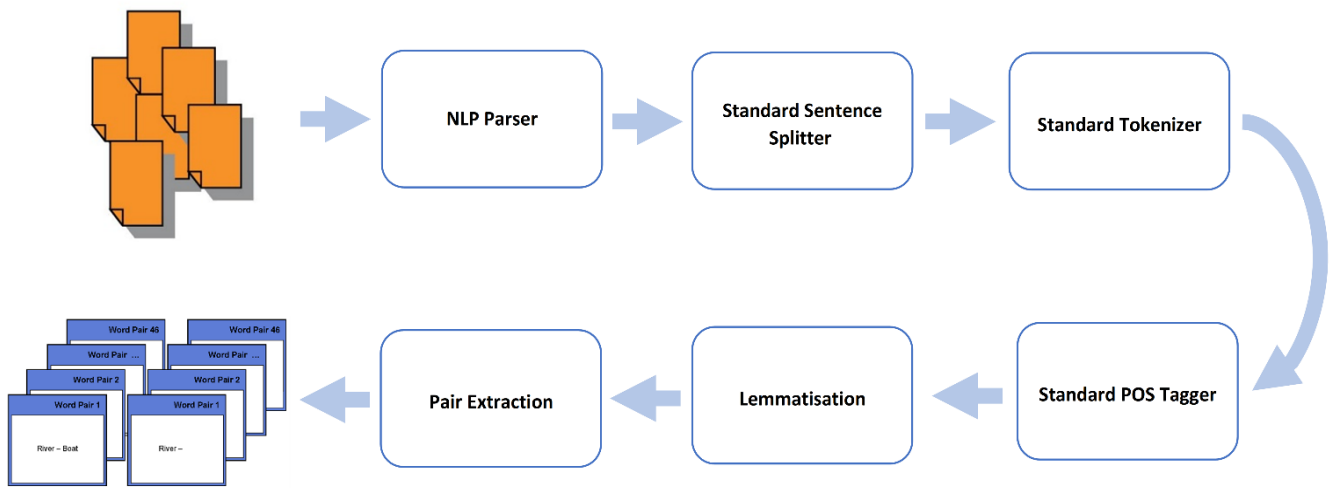

**Figure 1: NLP Components (redrawn from [3])**

Additionally, pair collection is smoothed to address data sparseness - meaning that the occurrence of pairs that disappear in the corpus is implied based on the frequency of other pairs. Finally, FCA is applied to the pairs as a formal framework. Figure 2 portrays the entire process of automatically constructing concept hierarchies from text corpora.

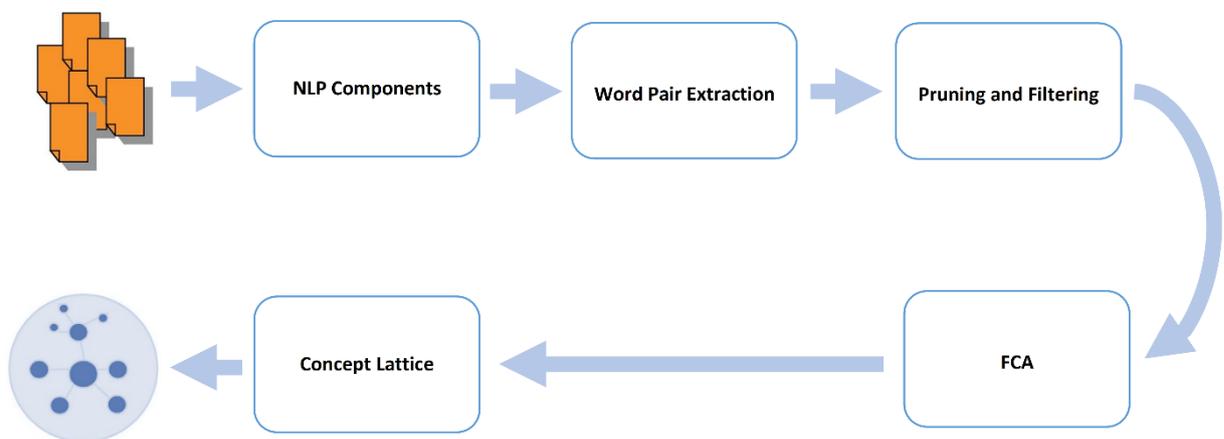

**Figure 2: A framework for automatic construction of concept hierarchies (adapted from [3])**

However, the extracted pairs may not always be exotic, and it might result in erroneous concept hierarchies to: derive the right and interesting pairs, and decrease the formal context size. Reducing the size of formal context may result in removing uninterested and erroneous pairs, taking less time to extract the concept lattice.

Despite the fact that Dias and Vieira in [9] tested a variety of techniques for concept lattice reduction, there are molecular approaches known about the reduced structures of concept lattice that have been generated. The authors of the same review classified the techniques for reducing concept lattice in the literature into three groups:

A. **Duplicate information remover:** In general, the simpler definition lattice of duplicate information remover approach has less characteristics and may be best suited to applications that involve direct contact with the user. This may happen whether by simple visual understanding of the lattice or exploitation of the current formal context's implications.

Table 1 summarises the key duplicate information remover techniques for idea lattice reduction. In each of the six dimensions, objects represent techniques and attributes represent the characteristics of a technique.

**Table 1: The key Duplicate information remover techniques for concept lattice reduction from literature**

| Techniques | Features | Literature work (sources) | Sum |
|---|---|---|---|
| Start-up point | S1 | [10] | 1 |
| | S2 | [11–17] | 6 |
| | S3 | [12] | 1 |
| | S4 | [13, 18–26] | 10 |
| Prior knowledge | B1 | - | 0 |
| | B2 | [10, 11, 20–26, 12–19] | 17 |
| Formal context change | C1 | [11] | 1 |
| | C2 | [10, 11, 20–26, 12–19] | 17 |
| | C3 | [12, 14, 23–26, 15–22] | 14 |
| | C4 | - | 0 |
| End-result lattice | R1 | [10, 11, 20–26, 12–19] | 17 |
| | R2 | - | 0 |
| | R3 | - | 0 |
| Index quality | Q1 | [10] | 1 |
| | Q2 | [10] | 1 |
| | Q3 | [10, 11, 20–26, 12–19] | 17 |
| Algorithmic paradigm | A1 | [11, 12, 23, 26, 13–15, 17–21] | 12 |
| | A2 | [10, 16, 22, 24, 25] | 5 |
| | A3 | [11, 12, 21–26, 13–20] | 16 |

Where:

Start-up point is represented as follows: Context=S1, Concepts=S2, Lattice=S3, FCA extension=S4.

Prior knowledge is represented as follows: It is used=B1, It is not used=B2.

Formal context change is represented as follows: Objects=C1, Attributes=C2, Incidences=C3.

End-result lattice is represented as follows: Isomorphic to sandard=R1, A subset of sandard=R2, Not isomorphic nor subset=R3.

Index quality is represented as follows: Fidelity losses=Q1, Representativeness losses=Q2, Descriptive losses=Q3.

Algorithmic paradigm is represented as follows: Exact=A1, Heuristic=A2, High complexity=A3.

B. **Simplifiers:** This approach is more applied than previous ones by transforming the area of the definitions into a smaller one. This feature creates a situation where the consistency of the resulting definition lattice is in contrast with the standard one. In general, the methods are used in a formal setting and have a degree of complexity that enables them to be used in a variety of formal contexts. However, the end-result concept lattice may vary greatly from the original and, as a result, may be of low quality. Table 2 summarises the key simplifier techniques for idea lattice reduction. In each of the six dimensions, artefacts represent techniques and attributes signify the features of Simplifier technique.

**Table 2: The key Simplifier techniques for concept lattice reduction from literature**

| Techniques | Features | Literature work (sources) | Sum |
|---|---|---|---|
| Start-up point | S1 | [27–32] | 6 |
|  | S2 | [33–35] | 3 |
|  | S3 | [33, 36] | 2 |
|  | S4 | [37] | 1 |
| Prior knowledge | B1 | [28, 31, 34–37] | 6 |
|  | B2 | [27, 29, 30, 32, 33] | 5 |
| Formal context change | C1 | [31] | 1 |
|  | C2 | [28, 30, 32, 37] | 4 |
|  | C3 | [27–36] | 10 |
|  | C4 | - | 0 |
| End-result lattice | R1 | [37] | 1 |
|  | R2 | [34] | 1 |
|  | R3 | [27–33, 35, 36] | 9 |
| Index quality | Q1 | [27–33, 35–37] | 10 |
|  | Q2 | [27, 28, 37, 29–36] | 11 |
|  | Q3 | [27–33, 35–37] | 10 |
| Algorithmic paradigm | A1 | [27–29, 33] | 4 |
|  | A2 | [30–32, 34–37] | 7 |
|  | A3 | - | 0 |

C. **Selector:** This approach works by pruning the space of concepts using a few primary parameters, such as adding an objective function to remove apparently irrelevant paths on the concept lattice. The construction of iceberg concept lattices, as suggested by [38], is a well-known technique in this field. This approach has the downside of causing essential formal structures to be overlooked. Furthermore, some strategies compile a list of all formal concepts and use a collection of parameters to pick the ones that are applicable [39–41]. Since the entire search space is searched, any technique based on this approach can be expensive. Table 3 summarises the key Selector strategies for concept lattice reduction. In each of the six dimensions, artefacts represent techniques and attributes signify the features of a technique.

**Table 3: The key Selector techniques for concept lattice reduction from literature**

| Techniques | Features | Literature work (sources) | Sum |
|---|---|---|---|
| Start-up point | S1 | [39, 42–49] | 9 |
| | S2 | [39–41, 50] | 4 |
| | S3 | [38, 40, 41, 50–56] | 10 |
| | S4 | - | 0 |
| Prior knowledge | B1 | [39, 41, 44, 45, 48, 54, 56] | 7 |
| | B2 | [38, 40, 53, 55, 42, 43, 46, 47, 49–52] | 12 |
| Formal context change | C1 | [38–53] | 16 |
| | C2 | [38–48, 50–53, 55, 56] | 17 |
| | C3 | [38–54] | 17 |
| | C4 | - | 0 |
| End-result lattice | R1 | [56] | 1 |
| | R2 | [38–48, 50–53] | 15 |
| | R3 | [49, 54, 55] | 3 |
| Index quality | Q1 | [38–56] | 19 |
| | Q2 | [38–56] | 19 |
| | Q3 | [38–56] | 19 |
| Algorithmic paradigm | A1 | [38, 40, 51–53, 55, 56, 42–49] | 15 |
| | A2 | [39, 41, 50, 55, 56] | 5 |
| | A3 | [39, 40, 50, 54] | 4 |

In the same study [9], a total of forty techniques were analysed and classified based on seven criteria, selected from among the major ones in the literature. The results were summarised in a formal context, and the analysis was carried out using FCA. All reduction strategies have been shown to change the occurrence relationship and trigger various levels of descriptive loss. The first half of the techniques build on the collection of principles by focusing solely on duplicate information remover techniques, while the second half is based on an expansion of the FCA. They use exact algorithms in three-quarters of the cases, while heuristic methods are used in one-quarter of the cases. The majority of simplifier techniques start with the formal sense, generate a non-isomorphic non-subset of the original concept lattice, and use exact algorithms; others, on the other hand, depend on historical knowledge to aid in the simplification process. Finally, selector techniques start with a formal context, rely on no prior knowledge, and use precise algorithms. Simplifier and selector methods, which can significantly minimise the area to be reached, clearly make the most significant reductions. Simplifier methods, on the other hand, are inherently harmful because they can dramatically alter the collection of formal definitions. The process leading to such adjustments must ensure that the model lattice's critical components are maintained. While selector strategies are interesting because they prune the concept space, traversing the space must be achieved in such a way that the appropriate concepts are reached.

According to the literature, there have been few works on consistency measurements of reduced definition lattices, such as the suggestion of [31]. Even though each reduction strategy has distinct characteristics and each method has distinct objectives, the knowledge embodied in the final design lattice has limitations that must be calculated. When comparing various approaches, mechanisms for measuring information loss are also important. What's required is a way to establish which data has been saved, removed, added, or modified, so that the reduction output can be measured and compared to other techniques. Further research into such quality measures is needed. Thus, despite reviewing the current framework for concept hierarchy construction from text and concept lattice size reduction, this study focuses on proposing a framework for deriving concept hierarchies taking advantage of deducing formal context size to produce meaningful lattice and concept hierarchy accordingly. The formal context could be significantly large because a large corpus created a formal context. Meanwhile, depending on the size and nature of the formal context data, the formal context may be a time-consuming operation. As a workaround, adaptive ECA * is used to implement a new paradigm that eliminates the incorrect, uninteresting pairs in the formal context and shrinks the formal context's size, making the term lattice less time-consuming. An experiment is then conducted to compare the results of the reduced definition lattice to the regular one. Resultantly, this study endeavours with substantial reduction concept lattice preserving the best quality of the end-result concept lattice. Incidentally, to the best of our understanding, the concept lattice reduction for deriving concept hierarchies from corpora, like in the approach presented in this paper has not been applied before.

## 3. Our proposed framework

As we mentioned earlier, this work aims to decrease the size of formal context as possible as it could remain the reduced concept lattice to the best quality. There are several methods for decreasing the size of a formal context. The used technique in this research for concept lattice reduction are ones that aim at reducing the sophistication of a concept lattice relating to interrelationships and magnitude, whereas retaining appropriate information. Reducing the size of concept lattice leads to derive less sophisticated concept hierarchies. Figure 3 illustrates the flow of producing the concept hierarchies using adaptive ECA* with the aid of WordNet.

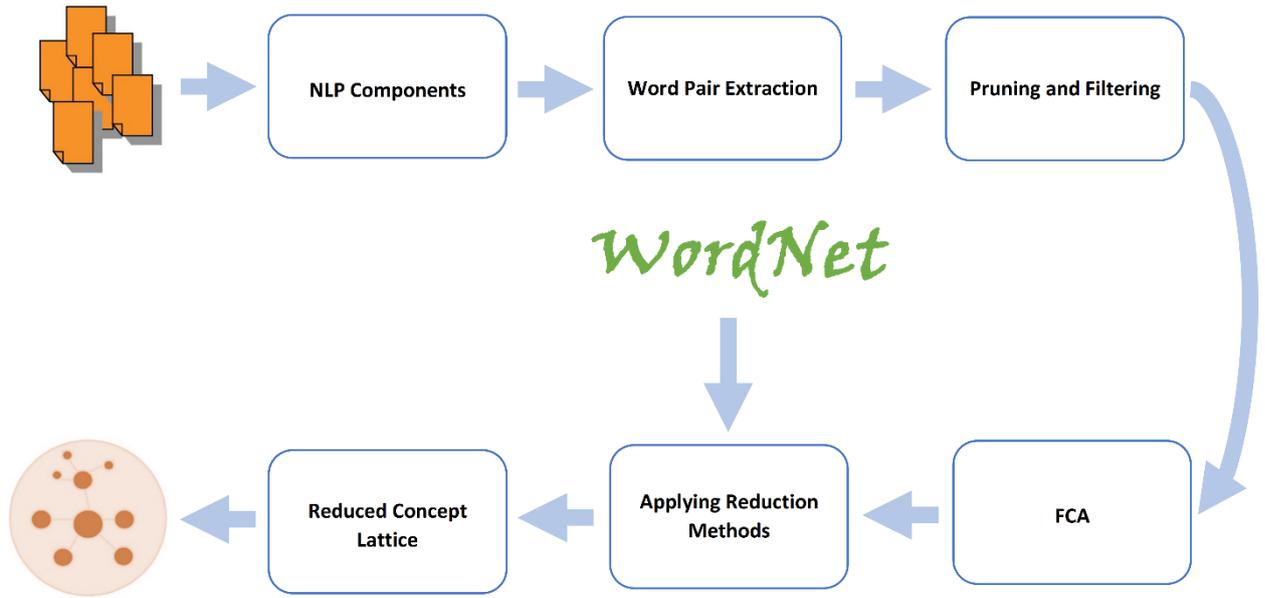

**Figure 3: A proposed framework of deriving concept hierarchies using adaptive ECA***

Since the introduced approach was based on multi-disciplinary techniques, it can, therefore, be highly effective [4]. After FCA applied to the pruned word pairs, a formal context is constructed. The formal context provides details about how attributes and objects should be related together. That means, formal context is a logical table that can be depicted by a triplet *(O, A, I)*, in which *I*, has a binary relationship between *O* and *A*. Elements of *O* are called objects that match the table rows. At the same time, elements of A are named attributes and match the table columns. Also, for $o \in O$ and $a \in A$, $(o, a) \in I$ indicates that object *o* has attribute *y* whereas $(o, a) \notin I$ indicates that *o* does not have *a*. For example. Table 4 presents logical objects and their attributes. The corresponding triplet *(O, A, I)* is given by $O = \{o_1, o_2, o_3, ..., o_n\}$, $A = \{a_1, a_2, a_3, ...., a_m\}$, and we have $(o_1, a_2) \in I$, $(o_2, a_3) \notin I$.

**Table 4: A formal context example**

|  | $A_1$ | $A_2$ | $A_3$ | $A_m$ |
|---|---|---|---|---|
| $O_1$ | x | x | x | $x_{1m}$ |
| $O_2$ | x | x | Nil | $x_{2m}$ |
| $O_3$ | Nil | x | x | $x_{3m}$ |
| $O_n$ | $x_{n1}$ | $x_{n2}$ | $x_{n3}$ | $x_{nm}$ |

Adaptive ECA* aims to reduce to the size of this table by applying the mut-over operator on its objects and attributes. For example, if $A_1$ is similar to $A_3$, a crossover is occurred on these attributes to generate $A_{13}$. The result of crossover can be a hypernym of both attributes based on WordNet. The similarity and relativity check are also applied to the objects *O*. Meanwhile, if $O_2$ and $O_3$ are related together, one of the attributes of both of them are mutated to find a common hypernym between them, and it can be presented as $O_{23}$. This relativity check is also applied to the pairs of attributes *A*. Table 5 depicts the result of Table 4 after applying adaptive ECA*.

**Table 5: Reduced formal context example after applying adaptive ECA***

|  | $A_{13}$ | $A_2$ | $A_m$ |
|---|---|---|---|

| $O_1$ | x | x | $x_{1m}$ |
|---|---|---|---|
| $O_{23}$ | x | x | $x_{2m}$ |
| $O_n$ | $x_{n1}$ | $x_{n2}$ | $x_{nm}$ |

In FCA, tables are commonly represented with logical attributes by triplets. We consider the *table (O, A, I)* instead of the *triplet (O, A, I)*. The goal of FCA is to obtain two outputs from the above table. The first one is a concept lattice, which is a half-organized collection of an individual object and attribute clusters. The second one consists of formulas which describe specific attribute dependences true in the table. For this study, the second goal can be achieved using ready software solutions, such as Concept Explorer.

### 3.1. Adaptive ECA*

ECA* consists of five components:

(1) Initialisation;

(2) Clustering I;

 (3) Mut-over;

(4) Clustering II;

(5) and Evaluation.

In this study, the operators of ECA* are adapted as follows: (1) Initialisation, (2) Construction I, (3) Mut-over, (4) Construction II, (5) and Evaluation. The ingredients of adaptive ECA* are depicted in Figure 4.

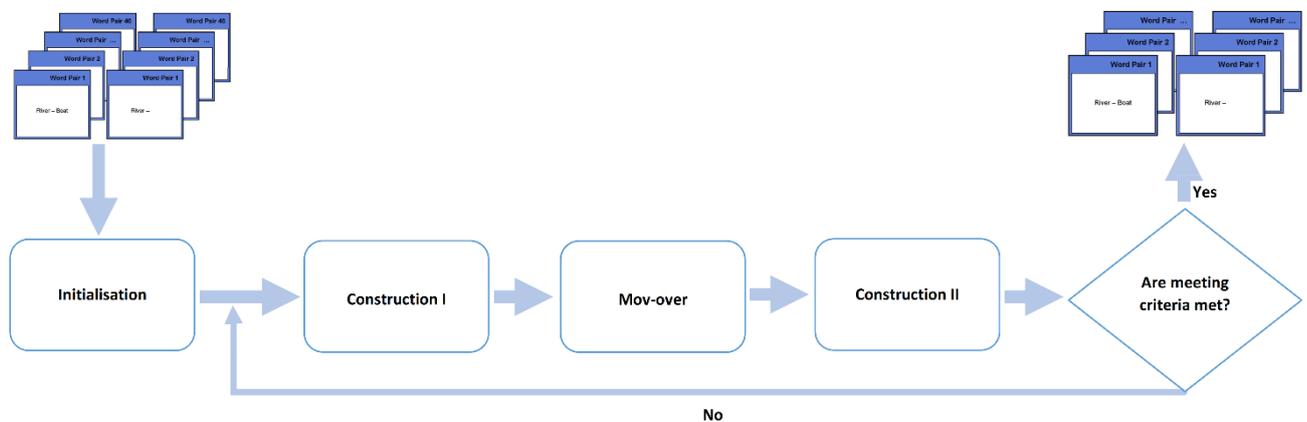

**Figure 4: Adaptive ECA***

The incorporation of stochastic and randomness processes is one of the ECA*'s advantages. The stochastic method in the ECA* has the benefit of striking a balance between exploring the search space and using the search space's learning process to zero in on local and global optima.

The use of heuristic operators is also credited with the ECAoutstanding *'s performance [6]. Meanwhile, three elements of our algorithm's mutation strategy have been altered [57]. To begin, the adaptive control parameter (F) is implemented by using Levy flight optimisation to balance the algorithm's exploitation and exploration. Second, in order to develop their learning abilities and find the best cluster centroids for clusters, the cluster centroids learn from historical cluster centroids (HI). Mut-over can also be used for mutation-crossover as a recombination strategy. Third, mut-overing with F and HI eliminates the issue of global and local optima that can occur with other algorithms such as K-means [58]. Thanks to this recombined evolutionary operator [57], our proposed algorithm is even more stable and robust. As a result, these techniques are effective in balancing the relationship between global and local optimality.

The components of adaptive ECA* are detailed as follows:

A. **Initialisation:** The hypernym and hyponym depths are initialised. Also, the number of iterating the algorithm is initialised.

B. **Construction I:** Consider a formal context as *(O, A, I)*, and each note structure of the context can be represented as *x*. Taking a set of the objects as $O = \{o_1, o_2, o_3, ..., o_n\}$, and a set of attributes as $A = \{a_1, a_2, a_3, ..., a_m\}$, objects can be presented as pairs $(o_i, o_{i+1})$, and also attributes can be represented as pairs $(a_j, a_{j+1})$. These representations are shown in Equation (1) and (2), respectively.

$$(o_i, o_j) = \sum_{i=1}^{n-1} \sum_{j=i+1}^{n} (o_i, o_j) \quad (1)$$

$$(a_i, a_j) = \sum_{i=1}^{m-1} \sum_{j=i+1}^{m} (a_i, a_j) \quad (2)$$

C. **Mut-over:** This operator comprised of the recombined technique of crossover and mutation. The operation of mut-over is depicted in Equation (3).

$$(oi, oj) = \begin{cases} o_i \text{ is similar to } o_j & \text{crossover (common hypernym)} \\ o_i \text{ relates to } o_j & \text{mutate both to find common hypernym)} \\ Otherwise & \end{cases} \quad (3)$$

1. Crossover: If two objects/attributes are similar to each other according to the WordNet, the crossover can be occurred on them to generate the most fitting general and hypernym synset for the close two objects/attributes in meaning. There are several methods to compute the similarities between the two terms. One of the most popular approaches is WordNet, which is used for measuring knowledge-based similarities between terms. Referring to [59], WordNet is a winner approach for similarity measurement tasks as it uses lexical word alignment. Meantime, WordNet is used in this work to determine how two terms are related together.

2. Mutation: If two objects/attributes are related to each other according to the WordNet, one or both of them can be mutated to find a common a hypernym between them. Their relationships can recognise the objects/attributes via the hypernym and hyponym depth.

3. If the objects/attributes are not either similar or relative to each other, no operations will be made on them.

D. **Construction II:** Deriving concept lattice from formal context. After that, find the lattice invariants to check to what extent the resulting lattice is isomorphic with the standard one.

E. **Evaluation:** evaluating the isomorphic level of the resulting lattice with the original one to terminate the algorithm or continue.

**3.2. Complexity issues**

Our algorithm presented above has several mechanisms. As a result, the mut-over strategy may be able to minimise the amount of time spent comparing and traversing the design lattice. While the time complexity of adaptive ECA* is the same as that of previous clustering evolutionary algorithms, the running time is decreased, allowing experimental findings to appear more rapidly. Resultantly, the adaptive ECA* has a bound of $O(|L| |G| |M|^3)$ of worst-case time complexity. The complexity is proportional to the number of times the mut-over operator is invoked with each extent of the definition lattice calling either crossover or mutation once. Furthermore, in different clustering dataset structures, pre-defining the number of initial iterations with compactness could cause an algorithm to execute with a linear time-complexity [60]. As a measure, the adaptive ECA's on-call complexity is roughly calculated as $O(|L| |G| |M|)$, and the overall complexity is $O(|L| |G| |M|^3)$.

**3.3. Adaptive ECA* parameter tuning**

Because the algorithms' clustering solutions can differ between runs, we run the adaptive ECA* 30 times per dataset to record the cluster quality for each run. The total number of iterations for each run is 50. We also keep track of the average results for each technique for the (30) times clustering solutions on each dataset problem. As part of mut-over strategy, the uniform crossover is exploited as it is one of the effective and efficient types of the operator in evolutionary algorithms for minimising common problems [61]. Finding the best value for pre-defined variables in ECA*, such as the number of social class ranks and the cluster density threshold, is also challenging. Choosing the right number of social class ranks will lead to the wrong number of clusters being found. Furthermore, ECA* fails to choose the appropriate cluster density threshold, which is critical for complex and multi-featured problems. Depending on the type of benchmarking issue, the cluster density threshold can be different. For example, a cluster density threshold of 0.001 might be optimal for one form of dataset but not for another. As a result, the cluster density threshold should be calculated based on the dataset's size and characteristics. Thus, we assume the initial values in Table 6 are the best for the dataset problems in this analysis.

Table 6: The pre-defined tuning parameters

| Adaptive ECA* parameters | Initial value |
| --- | --- |
| Cluster density ratio | 0.001 |
| Social class rank number | 2-10 |
| Iteration numbers | 50 |
| Number of executions | 30 |
| Crossover operator | Uniform crossover |

**4. Experimental analysis of the proposed framework**

This section consists of two parts: the datasets used in the experiment, and the detailed setup of the experiment.

To ignore the uninteresting and some false pairs in the formed formal context, we experiment to eliminate these word pairs and formal context size reduction accordingly. In the context of this experiment, we will attempt to address the following questions:

1. Can it decrease the size of the formal context without impacting the quality of the result?

2. Do we need an evolutionary algorithm with the aid of linguistic resources to decrease the size of the formal context keeping the quality of the resulting lattice?

3. Is adaptive ECA* with linguistic tools more effective in formal context size reduction than the current framework for reducing the concept of lattice size?

### 4.1. Datasets

It is somewhat helpful to use well-understood and commonly used standard datasets so that the findings can be quickly evaluated. Nevertheless, most of the corpora datasets are prepared for NLP tasks. Wikipedia is a good source of well-organised written text corpora with a wide range of expertise. These articles are available online freely and conveniently. Based on the number of English articles that exist in Wikipedia, we have randomly chosen several articles to be used as corpus datasets in this experiment. The sample size is determined from the confidence level, margin of error, population proportion, and population size. That means 385 articles are needed to achieve a 95 percent confidence level that the actual value is within ±5% of the calculated value. The parameters relating to the sample size of this study are mentioned in Table 7.

Table 7: Sample size calculation parameters

| Parameters | Value |
|---|---|
| Level of confidence | 95% |
| Error margin | 5% |
| Proportion of population | 50% |
| population size (number of English articles in Wikipedia) | 6090000 |

### 4.2. Experimental setup

The overall steps of our proposed framework, including the experiment, are as follows:

A. **Input data:** 385 text corpora are taken from Wikipedia that has different properties, such as length, and text cohort. The basic statistics of the free texts are calculated in Table 8.

Table 8: Basic statistics of the samples

| Length of data samples | Value |
|---|---|
| Mean | 556.842 |
| Median | 490.000 |
| Sum | 214384.000 |
| Max | 2821.000 |
| Min | 172.000 |
| STDV P | 298.225 |
| STDV S | 298.613 |
| STDEVA | 298.613 |

B. **Word pair extraction:** The corpora are labelled with part-of-speech tags before being scanned to generate a parsing tree for each sentence. The parser trees are then used to extract the dependencies. Following that, the pairs are lemmatized, the word pairs should be filtered and pruned, and the formal context-dependent on the word pairs should be constructed.

C. **FCA and concept lattice:** the formal contexts are formed from the word pairs to produce the concept lattices. Two categories of concept lattices are produced for evaluation as a result of implementing the experiment. The first concept of the lattice is formed directly from the text corpora without applying adaptive ECA*. In contrast, the latter one is constructed from the corpora after implementing the adaptive ECA* on it. These two concept lattices are labelled as follows:

1. Concept lattice 1: A concept lattice without applying any reduction method
2. Concept lattice 2: A concept lattice after applying adaptive ECA*.

Figure 5 illustrates the mechanism used to derive the two above concept lattices, and concept hierarchies as well.

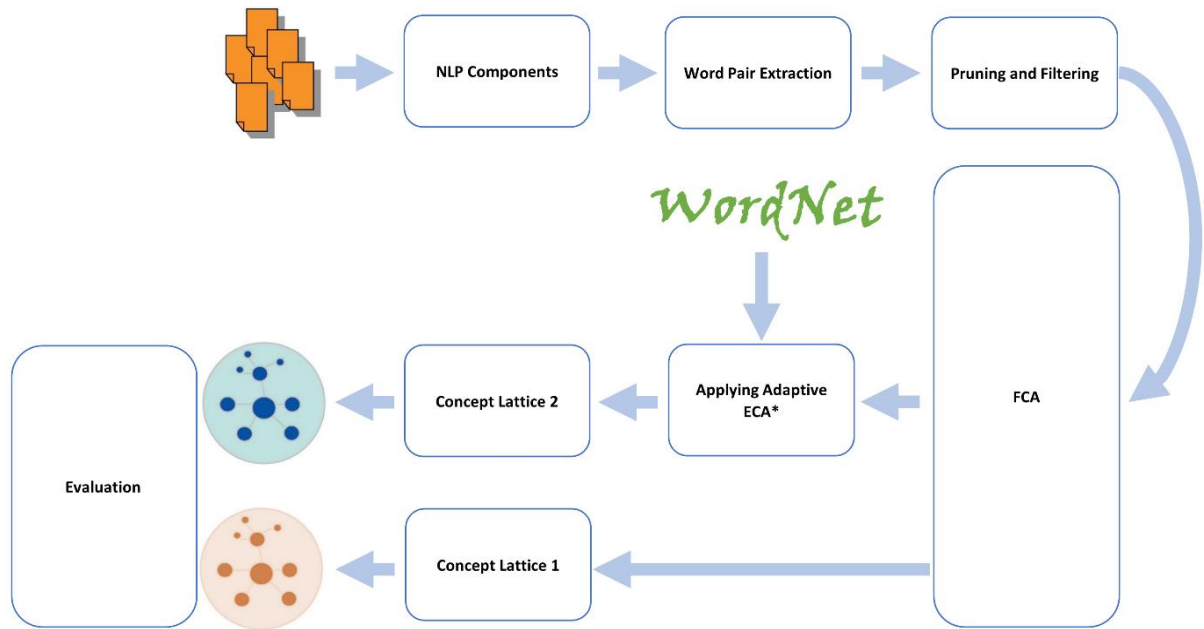

**Figure 5: A proposed framework for examining formal context size-reduction**

D. **Adaptive ECA*:** Once the concept lattices are generated, adaptive ECA* is applied to them. The parameters of constructing the concept lattice 1, and 2 are presented in Table 9.

**Table 9: parameters for deriving concept lattice 1, and 2**

| Parameters | Cocnept lattice 1 | Concept lattice 2 |
|---|---|---|
| Reduction method | - | Adaptive ECA* |
| Hypernym depth | - | 4 |
| Hyponym depth | - | 4 |
| Number of iterations | 1 | 30 |

E. **Concept lattice properties:** Following the derivation of the concept lattices, some statistical data from each is reported. The number of edges, height, and width of lattices are some of the statistical results.
F. **Evaluating the results:** At the last step, finding the relationships between the two lattices via isomorphic and homeomorphism.

### 4.3. Evaluation method

To evaluate our proposed framework compared to the standard framework for deriving concept hierarchies, we need to assess how well the concept lattices and concept hierarchies represent the given domain. Several methods are available to analyse lattice graphs. One possible method is to measure the similarity between the newly generated concept hierarchies to a given hierarchy for a specified domain. However, the difficultly is how to define the similarity and similarity measures between the hierarchy of concepts. Although a few studies are dealing with calculating the similarities between concept lattices, simple graphs, and concept graphs, It is unclear how to quantify and translate these similarities into concept hierarchies [62]. Accordingly, ontologies are evaluated with the pre-defined similarity procedures, and therefore the agreement on the task of modelling ontology is yielded from different subjects. Because concept lattice is a particular type of homomorphism of the structure, evaluating the concept lattices by finding a relationship between the lattice graphs is another evaluation method. In graph theory, if there is an isomorphism from the subdivision of the first graph to some subdivision of the second one, these two graphs are considered as homeomorphisms [21]. For example, if there is an isomorphism from the subdivision of graph $X$ to some subdivision of graph $Y$, graph $X$ and $Y$ are homeomorphisms. Furthermore, a subdivision of a graph in the theory of graphs can be defined as a graph formed by the subdivision of edges within a graph. In addition, subdividing an edge a with endpoints {x, y} produces a graph with one new vertex z and two new edges in place as {y, z} and {z, y}. This approach of evaluation seems to be more systematic and formal because finding two graphs is theoretically precise whether homeomorphism is or not. To determine whether the original concept lattice with the results ones are isomorphic or not, we need to find out common characteristics between them. Such characteristic is called concept lattice-invariant, which is maintained by isomorphism. The concept lattice-invariant, applied in several studies in the literature [9], include the number of concepts, the quantity of edges, degrees of the concepts, and length of the cycle. These invariants are also adapted for use in this work.

In this paper, these definitions express the concept of redundant information, isomorphism, and homeomorphism, respectively.

**Definition 1.** An object $o \in O$, and attribute $a \in A$, or occurrence $i \in I$ is regarded as If its removal or transformation results in a lattice that is isomorphic to to $\beta(O, A, I)$., it is redundant knowledge.

**Definition 2.** Formally, two lattice graphs $L_1 = (C_1, E_1)$ and $L_2 = (C_2, E_2)$ are isomorphic if there is a bijective function $f$ from $C_1$ to $C_2$ with the feature that $x$ and $y$ are adjacent in $L_1$ if $F(x)$ and $F(y)$ are adjacent in $L_2$ $\forall_{x,y} \in C_1$.

**Definition 3.** If a mapping $f: L_1 \rightarrow L_2$ between two concept lattices, these two lattices are called lattice homeomorphism if and only if they preserve supremum and infimum.

### 4.4. Result and analysis

To analyse the results of the experiment, we have evaluated the two concept lattices using the 385 text corpora based on concept lattice-invariant as it is represented by concept numbers, the quantity of edges, lattice height, and estimation of lattice width. Figure 6 presents the concept counts for concept lattice 1, and 2 for the used corpora in this study. It indicated that the concept numbers of lattice 2 is less than the ones in lattice 2. The mean impact of applying adaptive ECA* on the second concept lattice compared to its original one is the decrease of concept counts by 16%.

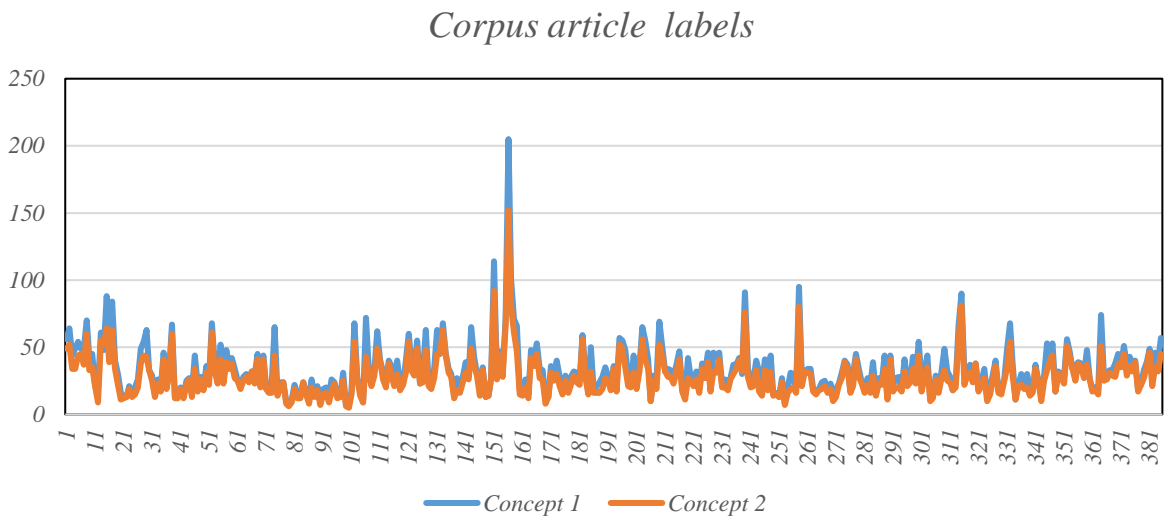

Figure 6: The concept counts for concept lattice 1, and 2 for the 385 corpus articles

Moreover, Figure 7 depicts the fluctuation of edge numbers for concept lattice 1 and 2 for the used corpora. The quantity of edges of lattice 2 is less than or equal to the ones in lattice 2 from corpus articles 1 to 385. On average, the concept lattice 2 is simplified by 16% in relation to their edges.

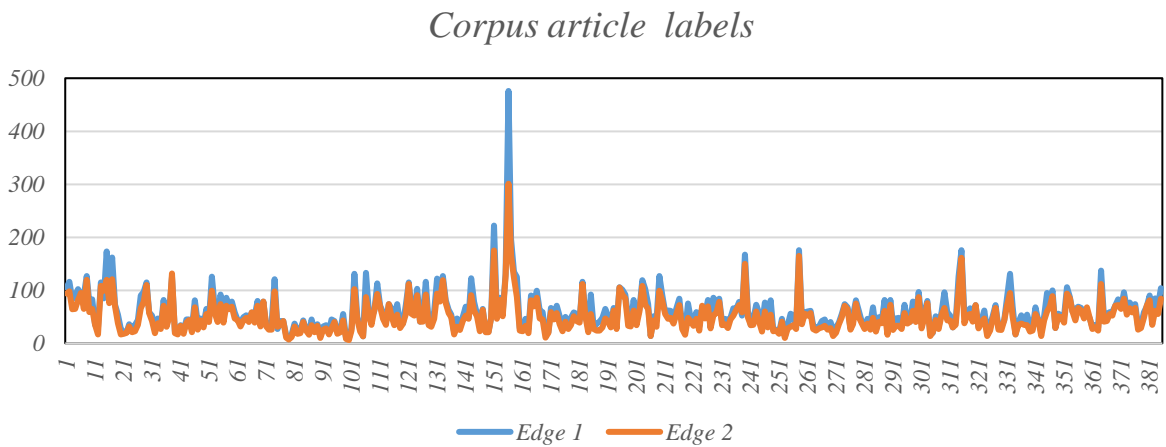

Figure 7: The fluctuation of edge counts for concept lattice 1, and 2 for the 385 corpus articles

For the heights of concept lattice 1, and 2, they have relatively the same lattice heights if we take the average heights of both lattices. Nevertheless, the heights of the resulting lattice (lattice 2) are changing from one text corpus into another one. In most cases, the heights of resulting lattice are equal to the original lattice. The heights of concept lattice 1, and 2 are depicted in Figure 8.

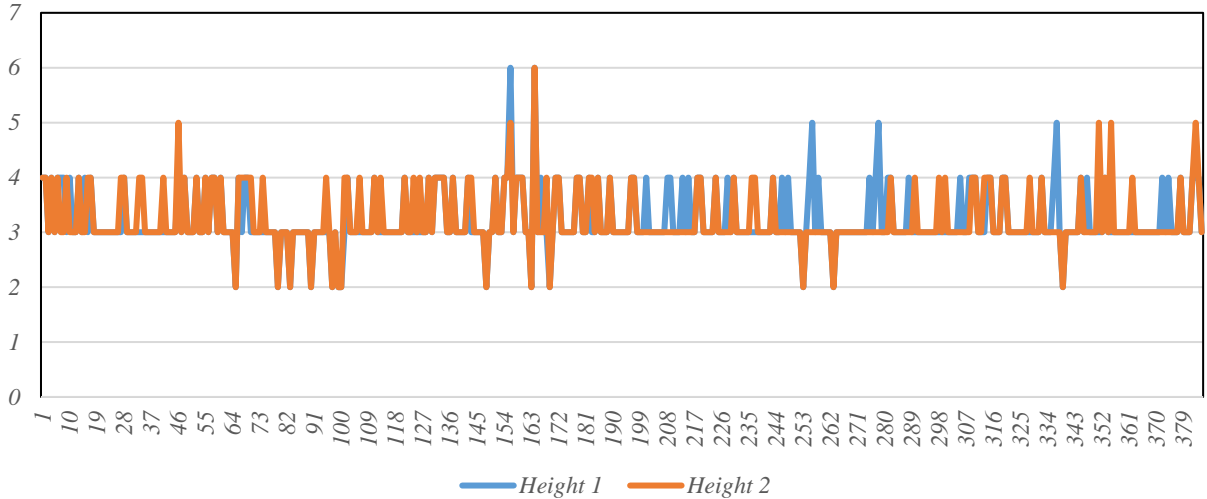

**Figure 8: The heights of lattice 1, and 2 for the 385 corpus articles**

Also, for the width of the original and resulting lattice, their widths are considered equal in several cases. In contrast, in some instances, there is a sharp fall in the resulting lattice's width. On the whole, the resulting lattice's width is reduced by one-third in comparison to the underlying lattice.

To further illustrate the results, the results of both concept lattices applied to the 385 datasets are aggregated in Table 10. In accordance with the aggregated values given in the table, the concept count, edge count, height, and width are substantially decreased in the resulting lattice compared to the concept lattice 1. As an instance, the average quantity of concepts and edges in the resulting lattice is decreased by 16% with similarly the same height of lattices. Meanwhile, the width of reduced concept lattices is considerably reduced compared to the original one. However, three factors might implicitly or explicitly affect the results of reduced concept lattices: the depth of hypernym and hyponym of the WordNet, the cohort, and length of text corpora.

**Table 10: Aggregated results of concepts and edges of concept lattice 1, and 2**

|  | Concept 1 | Concept 2 | Edge 1 | Edge 2 | Height 1 | Height 2 | Width 1 | Width 2 |
|---|---|---|---|---|---|---|---|---|
| **Mean** | 33.306 | 27.930 | 60.026 | 50.226 | 3.221 | 3.221 | [22.51,30.01] | [12.64, 24.61] |
| **Median** | 30.000 | 25.000 | 52.000 | 43 | 3.000 | 3.000 | [21,26] | [12, 21] |
| **Sum** | 12823.000 | 10753.000 | 23110 | 19337 | 1240 | 1240.000 | [8666,11554] | [4865, 9475] |
| **Max** | 205.000 | 152.000 | 476.000 | 301 | 6.000 | 6.000 | [86,200] | [51, 217] |
| **Min** | 7.000 | 5.000 | 10.000 | 6 | 2.000 | 2.000 | [4,5] | [3, 3] |
| **STDV P** | 18.697 | 14.994 | 38.559 | 30.882 | 0.535 | 0.530 | [11.26,18.47] | [5.64, 17.65] |
| **STDV S** | 18.722 | 15.014 | 38.609 | 30.922 | 0.536 | 0.531 | [11.27,18.5] | [5.65, 17.68] |
| **STDEVA** | 18.722 | 15.014 | 38.609 | 30.922 | 0.536 | 0.531 | [11.27,18.5] | [5.65, 17.68] |

Additionally, it can be seen that the end-result concept lattice is still isomorphic from the meanings of the basic concept of lattice, and the most simplified concepts are used in their formal context. There are, however, some invariants that do not occur in the concept lattice 2. In the end-result concept lattice, the objects and attributes of these non-existent concepts are the least used or may be uninteresting. Consequently, thanks to the common invariants between them, it can be seen that the decreased concept lattices have moderately the same quality of results as the standard concept lattice 1 (concept lattice 1). As a result, the original lattice is homeomorphism to the resulting lattice by 89%. That means, there is about 11% quality and meaning loss between the two lattices. This loss leads to the quality loss of the resulting concept hierarchy in contract with the original one.

## 5. Experimental comparison of adaptive ECA*

To show the advantage of ECA* in extracting and reducing the concept lattice and concept hierarchies in this study, we have used Java programming language to measure the execution time of the proposed algorithm against its competitive baseline algorithms (Fuzzy K-means, JBOS approach, AddIntent algorithm, and FastAddExtent). The rationale behind choosing these mentioned algorithms as competitive to adaptive ECA* is that these algorithms are the most recent and popular algorithms among others. Fuzzy K-means is one of the common algorithm, based on K-means, proposed by [30] for concept lattice reduction. This new approach was tested on two application areas, and the results show its robustness and efficient enough on practical applications. In another study, the junction based on objects similarity (JBOS) was introduced by [31]. This new approach was examined on datasets from the UCI Machine Learning Repository. The findings suggest that the JBOS method has the potential to minimise the size of a concept lattice. As per [63], AddIntent was compared with several efficient and popular counterpart algorithms (Norris, NextClosure, Bordat, Godin, and Nourine) for several types of datasets. The experimental results indicated that AddIntent is the best performed among its competitive algorithms. Furthermore, FastAddExtent can perform concept lattice faster than AddExtent in different fill ratios [64].

Also, this experiment was run on a personal computer with an idle Windows 10 64-bit operating system, CPU of Intel® Core™ i7-8550U @ 1.80 GHz (8 CPUs), ~2.0GHz, and 16 GB RAM. The datasets used in this experiment were created at random with three different densities (fill ratios):

- Low density: 10%
- Medium density: 25%
- High density: 50%

The number of objects in these random datasets is 100, while the number of attributes is variable, and each attribute may have a different number of objects.

On low density (10 percent fill ratio) random datasets, Figure 9 shows the runtime relation between the adaptive ECA* and counterpart algorithms (Fuzzy K-means, JBOS, AddIntent, and FastAddExtent). The number of attributes (|M|) increases steadily from 10 to 20,000, as seen in the diagram. When the |M| is insufficient, the adaptive ECA* deviates slightly from its competitors' methods. Nonetheless, as |M| increases, the adaptive ECA* gains an increasing advantage, and the gap widens.

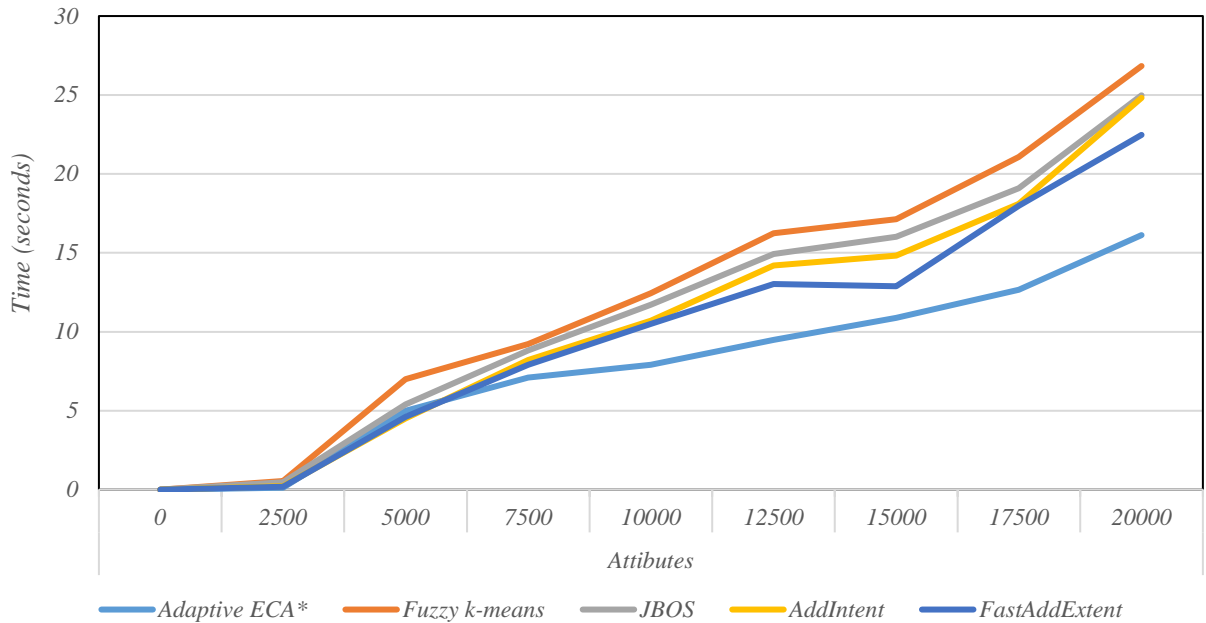

**Figure 9: Results of adaptive ECA* against its competitors for randomised datasets in 10% fill ratio (low density)**

Also, Figure 10 demonstrates the runtime comparison between the adaptive ECA* and its competitive algorithms medium density (25% fill ratio) random datasets. The number of attributes (|M|) increases steadily from 10 to 6000, as shown in the diagram. In contrast to Figure 9, the running time difference is greater. When there are a lot of attributes, the adaptive ECA* has a lot of advantages. Especially when |M| is around 1225 or less.

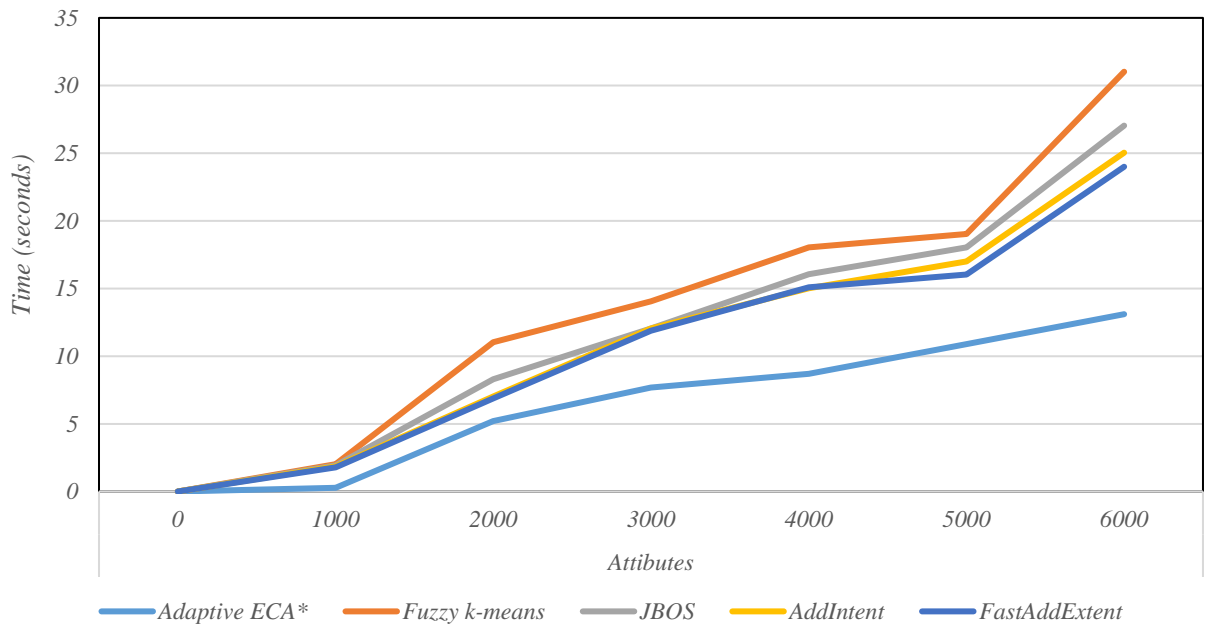

**Figure 10: Results of adaptive ECA* against its competitors for randomised datasets in 25% fill ratio (medium density)**

Figure 11 demonstrates the run comparison between adaptive ECA* and its counterpart algorithms on high density (50% fill ratio) random datasets. The number of attributes (|M|) steadily increases from 10 to 400, as seen in the diagram. The running time is increasingly increasing, as shown by the line chart obtained from the experiment. In contrast to Figures 9 and 10, the intersecting point was visible earlier. In the meantime, as compared to competing algorithms, the adaptive ECA* has clear advantages at each test stage.

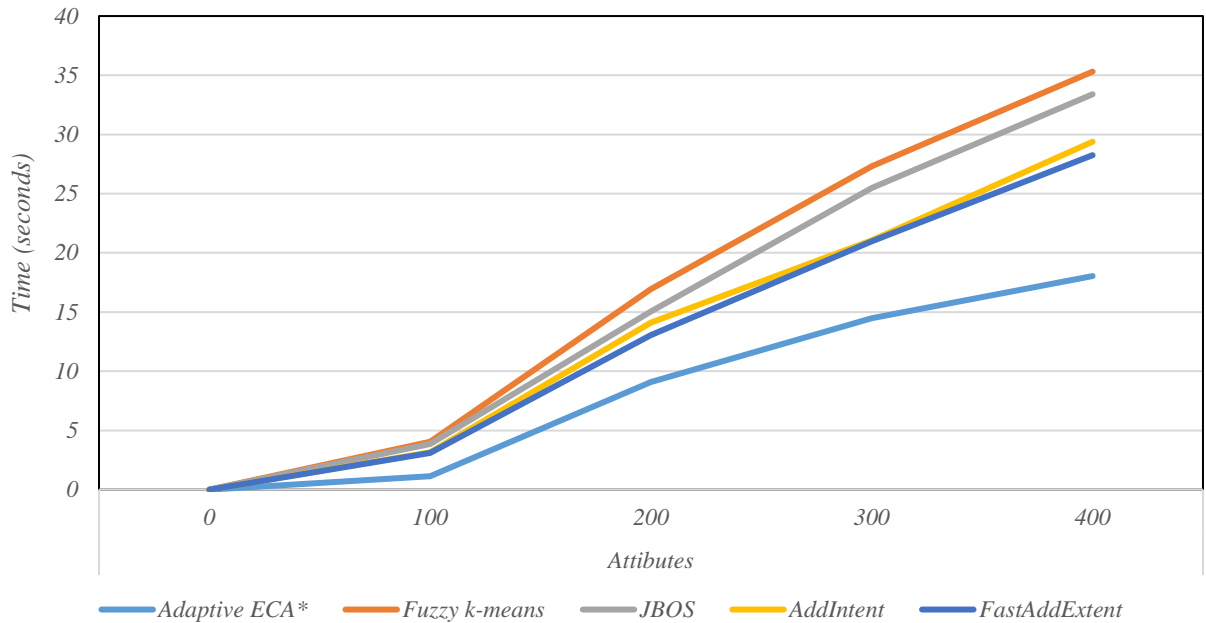

**Figure 11: Results of adaptive ECA* against its competitors for randomised datasets in 50% fill ratio (high density)**

## 6. Conclusions

In this study, the main contributions were proposing two forms of framing the concept hierarchy derivation: (i) The state-of-art framework was reviewed for forming concept hierarchies from free text using FCA; (ii) A novel framework was proposed using adaptive ECA* to remove the flawed and uninteresting pairs in the formal context and to decrease the size of the formal context, resulting in less time consuming when concept lattice derives from this. Later, an experiment was conducted to examine the results of the reduced concept lattice against the original lattice. Through the lens of the experiment and result analysis, the edn-result lattice is homeomorphism to the standard one with preserving the structural relationship between the two concept lattices. This similarity between the two lattices preserves the quality of resulting concept hierarchies by 89% in contrast to the basic one. The quality of resulting concept hierarchies is significantly promising, and there is information loss by 11% between the two concept hierarchies. Last, an experimental evaluation of adaptive ECA* was conducted against its counterpart approaches to measure their running time of on random datasets with different fill ratio. The results indicate that adaptive ECA* performs faster than other algorithms for the random dataset with three different densities (low, medium, and high).

This study is not out of shortcomings. First, assigning an adaptive value for hypernym and hyponym depths sometimes can improve the quality of the produced concept lattices. Second, polysemy problems also remain with this framework because it does not have the potential to include signals or indications with many related meanings. As the framework cannot sign the specific meaning of several related meanings, a word is usually regarded as different from homonymy, where the multiple meanings of a word can be unrelated or unrelated. Last but not least, classification as a way of identifying text bodies into categories or subcategories has been ignored as this can contribute to a better quality of results and facilitate a decision for the assignment of a proper value of hypernyms and hyponyms depths. Further researches could be carried out in the future to yield a better quality of the resulting lattice with less loss of meaning. The future scope of this work can be an investigation on initiating an adaptive depth of hypernym and hyponym in the mut-over operator of adaptive ECA*. Along with this, adaptive ECA* can be used with a frequency-based technique interchangeably as a threshold to eliminate the least occurring pairs in the formal context. Furthermore, an adaptive version of ECA* could be utilised for wound image clustering and analysis in telemedicine and patient monitoring [65, 66]. Meanwhile, an improved ECA* could be also used in practical and engineering problems [67], library administration [68], e-government services [69], and multi-dimensional database systems [70], web science [71], and the semantic Web [72]. On the other hand, the proposed frameworks can be adopted to learn concept hierarchies from other Latin alphabet-based text corpora.

**Acknowledgements**

The authors appreciate the referee's valuable and profound comments. Based on their feedback, the technical content of this paper has been greatly improved.

**Compliance with Ethical Standards**

**Funding Sources:** Funding is not received.

**Conflict of interest:** There are no declaration of conflict of interest the authors.